\documentclass[10pt,twocolumn,letterpaper]{article}

\usepackage{cvpr}              
\usepackage{booktabs}    
\usepackage{graphicx}    
\usepackage{xcolor}      
\usepackage{amssymb}
\usepackage{graphicx} 
\usepackage{pifont}
\usepackage{makecell}
\usepackage{latexsym}
\usepackage{multirow}
\definecolor{cvprblue}{rgb}{0.21,0.49,0.74}
\usepackage[pagebackref,breaklinks,colorlinks,allcolors=cvprblue]{hyperref}

\def\paperID{1} 
\def\confName{CVPR}
\def\confYear{2025}

\title{Benchmarking Multi-modal Semantic Segmentation under Sensor Failures: Missing and Noisy Modality Robustness}

\author{  Chenfei Liao\textsuperscript{1,}\thanks{These authors have equal contributions.}\quad Kaiyu Lei\textsuperscript{2,}\footnotemark[1]\quad Xu Zheng\textsuperscript{1,3,}\thanks{Project lead.}\quad Junha Moon\textsuperscript{1}\quad Zhixiong Wang\textsuperscript{1}\\ Yixuan Wang\textsuperscript{1}\quad Danda Pani Paudel\textsuperscript{3}\quad Luc Van Gool\textsuperscript{3}\quad Xuming Hu\textsuperscript{1,4}\thanks{Corresponding author.} \\
  \\
  \textsuperscript{1}HKUST(GZ)\quad \textsuperscript{2}XJTU\quad \textsuperscript{3}INSAIT, Sofia University “St. Kliment Ohridski”,
  \textsuperscript{4}CSE, HKUST \quad
  \\%
}
\begin{document}
\maketitle
\begin{abstract}

Multi-modal semantic segmentation (MMSS) addresses the limitations of single-modality data by integrating complementary information across modalities. Despite notable progress, a significant gap persists between research and real-world deployment due to variability and uncertainty in multi-modal data quality. Robustness has thus become essential for practical MMSS applications. However, the absence of standardized benchmarks for evaluating robustness hinders further advancement.
To address this, we first survey existing MMSS literature and categorize representative methods to provide a structured overview. We then introduce a robustness benchmark that evaluates MMSS models under three scenarios: Entire-Missing Modality (EMM), Random-Missing Modality (RMM), and Noisy Modality (NM). From a probabilistic standpoint, we model modality failure under two conditions: (1) all damaged combinations are equally probable; (2) each modality fails independently following a Bernoulli distribution. Based on these, we propose four metrics—$mIoU^{Avg}_{EMM}$, $mIoU^{E}_{EMM}$, $mIoU^{Avg}_{RMM}$, and $mIoU^{E}_{RMM}$—to assess model robustness under EMM and RMM. This work provides the first dedicated benchmark for MMSS robustness, offering new insights and tools to advance the field.
Source code is available at \href{https://github.com/Chenfei-Liao/Multi-Modal-Semantic-Segmentation-Robustness-Benchmark}{https://github.com/Chenfei-Liao/Multi-Modal-Semantic-Segmentation-Robustness-Benchmark}.
\end{abstract}
\section{Introduction}
\begin{figure}[t!]
    \centering
    \includegraphics[width=0.75\linewidth]{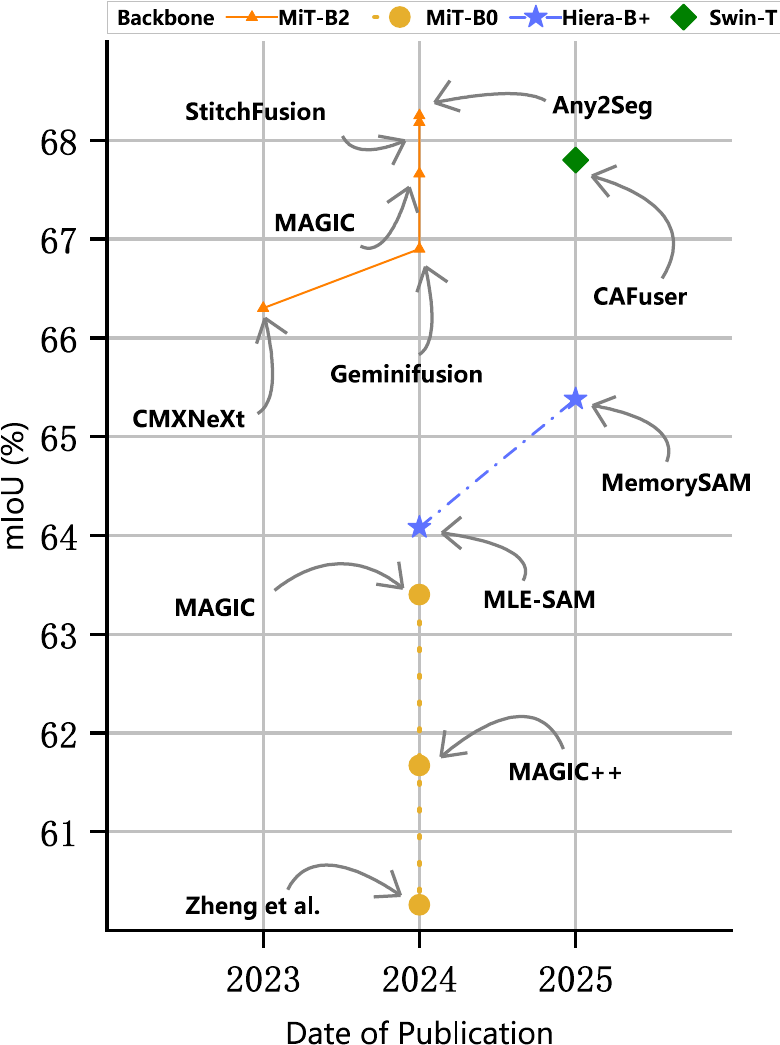}
    \caption{History of MMSS methods. }
    \label{fig:cover}
    \vspace{-12pt}
\end{figure}
Multi-modal semantic segmentation has emerged as a critical task in computer vision, leveraging diverse sensor inputs to produce more accurate and reliable pixel-wise classification results~\cite{mo2022review,zhang2021deep}. By fusing information from complementary modalities such as RGB, depth, LiDAR, thermal, and event, multi-modal systems  are capable of overcoming the limitations inherent to single-modality approaches~\cite{feng2020deep,song2023rgb,wang2021brief,zheng2023deep}. Such integration is particularly beneficial in autonomous driving, robotics, and surveillance applications, where harsh environmental challenges like low-light and adverse weather conditions may severely degrade the performance of individual sensors~\cite{zhang2024generative, liu2023improving, liu2021benchmarking, wang2025umsss}.

Despite these advantages, deploying multi-modal systems in real-world environments presents critical robustness challenges that are often underexplored in existing research~\cite{zhang2024multimodal}. In practice, sensor data may be incomplete, degraded, or entirely unavailable. For example, RGB cameras may struggle in low-light or foggy conditions; LiDAR sensors can produce sparse or noisy point clouds in heavy rain; thermal cameras are sensitive to ambient temperature variations; and event cameras may fail in low-contrast scenes~\cite{deng2021feanet,li2022rgb,wu2022complementarity,kachole2023bimodal,xie2024eisnet,yao2024sam,zhang2021issafe}. These scenarios illustrate the pressing need to evaluate and improve the robustness of multi-modal semantic segmentation (MMSS) models under real-world conditions.

To systematically address these issues, robustness in MMSS can be classified into three representative failure scenarios. First, Entire-Missing Modality (EMM) refers to the complete loss of a sensor's input, requiring the model to perform without that modality~\cite{10588722,zheng2024magic++,zheng2024learning}. Second, Random-Missing Modality (RMM) captures intermittent or partial sensor failures that cause unpredictable data absence. Third, Noisy Modality (NM) describes cases where sensors continue to provide input, but the data is degraded or corrupted due to environmental or hardware factors~\cite{zheng2024centering,zhu2024customize}. While some studies have explored EMM and NM, RMM—despite its high practical relevance—remains largely underexamined in the current literature.

In this paper, we present a comprehensive benchmark to systematically evaluate modality robustness in Multi-Modal Semantic Segmentation (MMSS). Our study begins by surveying and categorizing existing MMSS methods according to their architectural design principles and fusion strategies~\cite{zhang2023delivering,zhang2021deep}. We identify three primary approaches: (1) RGB-centric methods that use other modalities as supplementary information~\cite{zhang2023delivering}; (2) Equal-contribution methods that treat all modalities with uniform importance~\cite{liang2022multimodal}; (3) Adaptive-selection methods that dynamically determine modality contributions~\cite{li2024stitchfusion,zheng2024centering}. We then evaluate these methods under our three robustness scenarios (EMM, RMM, and NM) using the DELIVER dataset~\cite{zhang2023delivering}, which provides multi-modal acquired data under diverse environmental conditions. Our systematic evaluation reveals strengths and weaknesses in current approaches and illuminates promising directions for more robust systems.

Our contributions can be concluded as follows: 
(I) We comprehensively collect the works related to the multi-model semantic segmentation task and categorize them systematically.
(II) We build a robust benchmark for the multi-model semantic segmentation task, including 3 scenarios: EMM, RMM, and NM.
(III) From a statistical point of view, we propose new metrics to evaluate the performance of the model in both EMM and RMM cases.

\section{Related work}

\begin{table}[!t]
    \centering
    \caption{Comparison of modality robustness across MMSS methods. RMM: Random-Missing Modality; EMM: Entire-Missing Modality; NM: Noisy Modality.}
    \vspace{-6pt}
    \label{comparison}
    \small
    \resizebox{0.95\linewidth}{!}{
    \begin{tabular}{l|l|c|c|c}
    \toprule
    \textbf{Work} & \textbf{Publication} & \textbf{RMM} & \textbf{EMM} &\textbf{NM}\\ \midrule
    MCubesNet~\cite{liang2022multimodal} & CVPR2022 & \textcolor{red}{\ding{55}} & \textcolor{red}{\ding{55}} &\textcolor{red}{\ding{55}} \\ \midrule
    TokenFusion~\cite{wang2022multimodal} & CVPR2022 & \textcolor{red}{\ding{55}} & \textcolor{red}{\ding{55}} &\textcolor{red}{\ding{55}} \\ \midrule
    CMXNeXt~\cite{zhang2023delivering} & CVPR2023 & \textcolor{red}{\ding{55}} & \textcolor{red}{\ding{55}} &\textcolor{red}{\ding{55}} \\ \midrule
    GeminiFusion~\cite{jiageminifusion} & ICML2024 & \textcolor{red}{\ding{55}} & \textcolor{red}{\ding{55}} & \textcolor{red}{\ding{55}} \\ \midrule
    MAGIC~\cite{zheng2024centering}& ECCV2024 & \textcolor{red}{\ding{55}} & \textcolor{green}{\ding{51}} & \textcolor{green}{\ding{51}} \\ \midrule
    Any2Seg~\cite{zheng2024learning} & ECCV2024 & \textcolor{red}{\ding{55}} & \textcolor{green}{\ding{51}} & \textcolor{green}{\ding{51}} \\ \midrule
    FPT~\cite{10588722} & IV2024 & \textcolor{red}{\ding{55}} & \textcolor{green}{\ding{51}} & \textcolor{green}{\ding{51}} \\ \midrule
    MAGIC++~\cite{zheng2024magic++}& Arxiv2024 & \textcolor{red}{\ding{55}} & \textcolor{green}{\ding{51}} & \textcolor{green}{\ding{51}} \\ \midrule
    MLE-SAM~\cite{zhu2024customize} & Arxiv2024 & \textcolor{red}{\ding{55}} & \textcolor{green}{\ding{51}} & \textcolor{green}{\ding{51}} \\ \midrule
    AnySeg~\cite{zheng2024learning1} &Arxiv2024 & \textcolor{red}{\ding{55}}& \textcolor{green}{\ding{51}}& \textcolor{red}{\ding{55}} \\ \midrule
    StitchFusion~\cite{li2024stitchfusion} & Arxiv2024 &\textcolor{red}{\ding{55}} &\textcolor{red}{\ding{55}} &\textcolor{red}{\ding{55}} \\ \midrule
    CAFuser~\cite{brodermann2025cafuser} & RAL2025 &\textcolor{red}{\ding{55}} &\textcolor{red}{\ding{55}} &\textcolor{red}{\ding{55}}\\ \midrule
    MemorySAM~\cite{liao2025memorysam} & Arxiv2025 &\textcolor{red}{\ding{55}} &\textcolor{red}{\ding{55}} &\textcolor{red}{\ding{55}}\\
    \bottomrule
    \end{tabular}
    }
    \vspace{-12pt}
\end{table}
\subsection{Related Surveys and Benchmarks}
The development of intelligent vision sensors has sparked extensive research on multi-modal semantic segmentation, resulting in several surveys and benchmarks in this field. 
\textbf{The Surveys} on multi-modal semantic segmentation provide a comprehensive summary of existing methods, covering both bi-modal semantic segmentation and multi-modal semantic segmentation. For instance,~\cite{wang2021brief, song2023rgb, zheng2023deep} focus on the various modality combinations such as RGB-Depth, RGB-Thermal, RGB-Event, and so on. As to the multi-modal semantic segmentation,~\cite{zhang2021deep} offers an in-depth review of fusion strategies for handling multi-modal data. Moreover,~\cite{feng2020deep} concentrates on autonomous driving scenarios, providing insights from the perspective of real-world applications. 
\textbf{The Benchmarks} offer a fair comparison of advanced methods, providing effective performance measurements. Currently, DELIVER~\cite{zhang2023delivering}, MUSES~\cite{brodermann2025muses}, and MCubes~\cite{liang2022multimodal} are the most common benchmarks for multi-modal semantic segmentation, enabling researchers to test their methods under various modality combinations. 

The relevant surveys and benchmarks lay their emphasis on the accuracy of semantic segmentation models with multi-modal settings. While, in real-world applications, the multi-sensor systems will not permanently be as perfect as assumed. Besides accuracy, the robustness of multi-modal semantic segmentation models also plays a crucial role to the entire system, especially when sensors are disturbed or malfunctioning. \textit{\textbf{However, building a robustness benchmark of multi-modal semantic segmentation for multi-sensor systems remains a research gap. Our work is the first attempt to address this gap, hoping to bring new insights to the multi-modal semantic segmentation task.}}

\subsection{Multi-modal Semantic Segmentation}
\label{MMSS}
As a vital task in the computer vision field, semantic segmentation aims to allocate a class for each pixel~\cite{mo2022review}. Due to the lack of multi-modal datasets, previous research mainly focuses on the uni-modality~\cite{segformer, isnet, enet, fastscnn, bisenet, bisenetv2, stdc, pp-liteseg, rtformer, ddrnet} or bi-modality such as RGB-Depth~\cite{chen2020bi, cao2021shapeconv, chen2021spatial, seichter2021efficient, yin2023dformer}, RGB-Thermal~\cite{deng2021feanet, li2022rgb, wu2022complementarity, zhou2023dbcnet, zhao2023mitigating}, RGB-Event~\cite{zhang2021issafe, xie2024eisnet, kachole2023bimodal, yao2024sam}, and so on. With the development of vision sensor techniques and relevant multi-modal datasets, several multi-modal semantic segmentation models~\cite{liang2022multimodal, zhang2023delivering, zheng2024centering, li2024stitchfusion} are proposed to better cope with real-world requirements. which is shown in Fig.\ref{fig:cover}.
From the perspective of modality contribution, existing multi-modal semantic segmentation models can be classified into 3 types. \textcircled{1} Take RGB as the main contributor~\cite{zhang2023delivering}. CMXNet~\cite{zhang2023delivering} designs a Self-Query Hub to choose informative features from other modalities, which serve as supplements to the primary RGB information. \textcircled{2} Take all modalities as equal contributors~\cite{jiageminifusion, zhu2024customize, li2024stitchfusion, wang2022multimodal, zhu2024customize, brodermann2025cafuser}.MCubeSNet~\cite{liang2022multimodal} fuses multi-level features of different modalities in a concatenation way. \textcircled{3} Find the main contributor adaptively~\cite{zheng2024centering,liao2025memorysam, zheng2024magic++, zheng2024learning, zheng2024learning1}. Most representatively, MAGIC~\cite{zheng2024centering} determines the main contributor based on the similarity between each modality's features and the aggregated features. 
From these three design paradigms, several initial conjectures can be given. Firstly, Type\textcircled{1} is supposed to be more sensitive to the RGB modality. Its robustness relies on the stability of the RGB camera, which means the vulnerability of the model to RGB-in-friendly environments such as at night, cloudy, and so on. Secondly, Type\textcircled{2} is supposed to be moderately robust. Such models will not be greatly degraded by the interference or absence of a certain modality. Thirdly, Type\textcircled{3} is supposed to be equipped with the best modality robustness. The adaptive selection of the main contributor decreases the influence of the bad modality on the entire model. The experiments and further discussions are shown in Section \ref{experiments} and Section \ref{Discussion}.

\subsection{Modality Robustness} 
Research endeavors have focused on designing robust multi-modal frameworks to handle the modality-incomplete data~\cite{zheng2024centering, zheng2024learning1, 10588722, zheng2024learning, zhu2024customize, zheng2024magic++}. Current works mainly focus on the entire-missing modality (EMM) condition and noisy modality (NM) condition as shown in Table \ref{comparison}, which clearly compares how different methods emphasize modality robustness. In more detail, Any2Seg~\cite{zheng2024learning} attempts to utilize knowledge distillation from MVLMs to solve modality-agnostic segmentation. MAGIC~\cite{zheng2024centering} and MAGIC++~\cite{zheng2024magic++} attempt to evaluate each modality’s contribution to facilitate efficient cross-modal fusion, especially when faced with 
EMM and NM.
However, as in Table~\ref{comparison}, each study targets different aspects of modality robustness. Thus, we hope to establish a benchmark that evaluates how existing methods perform when dealing with the modality robustness problem. In addition, while most current methods focus on EMM and NM, they often skip random-missing modality (RMM) scenarios, which are closer to the real-world application. Our proposed benchmark will include RMM, EMM, and NM conditions along with all the existing open-source multi-modal semantic segmentation methods, attempting to bring new insights for future works.
\begin{figure*}[t!]
    \centering
    \includegraphics[width=0.78\linewidth]{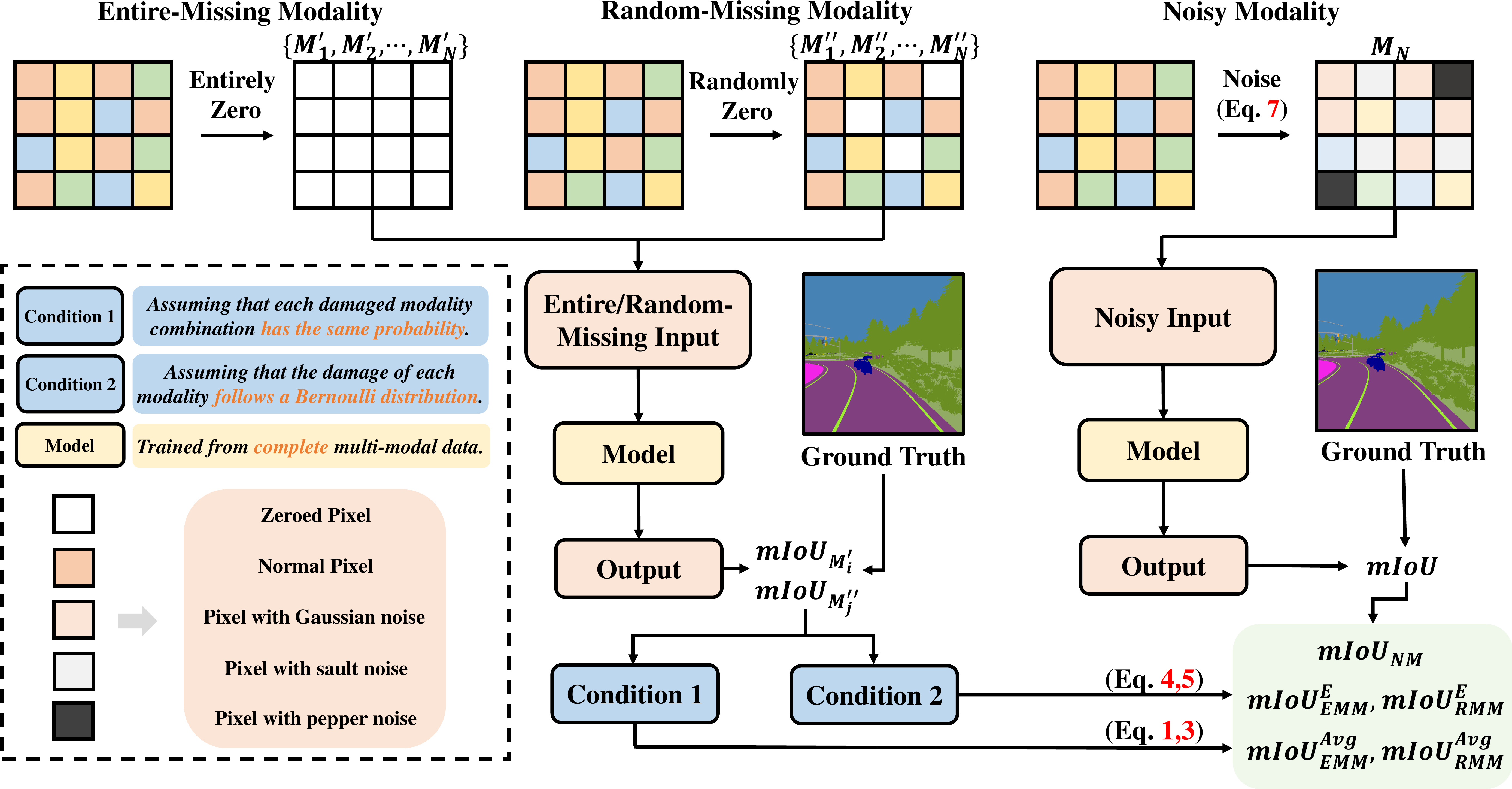}
    \vspace{-6pt}
    \caption{Framework of our multi-modal semantic segmentation robustness benchmark.}
    \vspace{-12pt}
    \label{fig:enter-label}
\end{figure*}
\section{Our Benchmarks}
\subsection{Evaluation Datasets}
\label{Datasets}
In this research, we use the DELIVER~\cite{zhang2023delivering} multi-modal dataset to evaluate semantic segmentation models. The dataset includes Depth, LiDAR, Multiple Views, Events, and RGB images, captured under five weather conditions (cloudy, foggy, night-time, rainy, and sunny). Each weather condition also includes five corner cases: Motion Blur (MB), Over-Exposure (OE), Under-Exposure (UE), LiDAR-Jitter (LJ), and Event Low-resolution (EL), which reflect real-world sensor performance challenges.

The DELIVER~\cite{zhang2023delivering} dataset was generated using the CARLA simulator~\cite{dosovitskiy2017carla}, a widely used open-source platform for autonomous driving research. It is specifically designed to simulate diverse and dynamic urban driving environments, ensuring that models can be evaluated under a wide range of practical, real-world conditions. The dataset provides six views per sample, each containing four modalities and two types of labels (semantic and instance segmentation), creating a rich multi-sensor setup. This makes the dataset ideal for evaluating model robustness under the context of autonomous driving, where conditions such as varying weather, sensor degradation, and complex environments are common.
With 25 semantic classes, the DELIVER\cite{zhang2023delivering} dataset enables a comprehensive evaluation of model performance across diverse segmentation tasks. Its carefully curated collection of data, encompassing diverse environmental challenges, serves as a valuable benchmark for assessing how well segmentation models generalize to real-world autonomous driving scenarios. The dataset's unique combination of multi-modal data and real-world simulations makes it particularly well-suited for evaluating the performance and robustness of models under conditions that closely resemble those encountered in actual deployments.

\subsection{Evaluation Methods}
\subsubsection{Entire-Missing Modality}
\label{EMM}
In real-world applications, multi-modal systems often face sensor damage. The most severe case is a complete failure of the sensors, which means that the corresponding modal data is entirely zeroed out. For instance, when the depth camera is entirely broken in an intelligent vehicle, all depth data becomes unavailable. To simulate this situation, we set the missing modalities’ data to zero and then make the entire data go through the trained model based on the full modality combination. 
Let the complete modality set as \( M = \{ m_1, m_2, m_3, \ldots, m_n \} \) and represent the missing modality combinations as \( \{ M_1', M_2', \dots, M_N' \} \subseteq M \), where each \( M'_i \) is a combination obtained by removing certain modalities from the complete modality combination \( M \), and \( N \) denotes the number of different \( M' \). 

To evaluate the model's performance under missing modalities, we first define \( mIoU_{EMM} \) as the mean intersection over union (IoU) for the missing modality combinations, as shown in Eq.~\ref{Eua1}, 
where \( mIoU^{Avg}_{M'_i} \) represents the validation mIoU for the corresponding missing modality combination \( M'_i \), using the model weights trained on the complete modality combination. 

\begin{equation}
    mIoU_{EMM}^{Avg} =\frac{1}{N} \sum_{i=1}^{N} mIoU_{M'_i}. 
    \label{Eua1}
\end{equation}

Furthermore, to consider the case of equal damage probabilities across all modalities, we further define \( mIoU_{EMM}^{E} \) as the expected mIoU under the assumption of equal damage probabilities for each individual modality. 
Assuming that the damage of each modality follows a Bernoulli distribution with a probability \( p \), we can represent the expected mIoU for all combinations of missing modalities. The probability of a specific combination of \( k \) modalities being damaged while the remaining \( n-k \) modalities are not damaged can be expressed using the binomial distribution. Specifically, for a combination \( M'_i \) that includes \( k \) damaged modalities, the probability $P( M^{'k}_i )$ is given by Eq.~\ref{Eua2}.  \( mIoU_{EMM}^{E} \) can then be calculated as Eq.~\ref{Eua3}, where \( M^{'k}_i \) is a combination obtained by removing \( k \) modalities from the complete modality combination \( M \), and \( \binom{n}{k} \) denotes the number of ways to choose \( k \) modalities from \( n \). The example of the R-D-E-L modality combination is shown in Table \ref{tab:modality_probabilities}.
\begin{equation}
P_p( M^{'k}_i) = p^k \cdot (1-p)^{n-k},
    \label{Eua2}
\end{equation}
\begin{equation}
mIoU^{E}_{EMM}(p) = \sum_{k=0}^{n-1} \sum_{i=1}^{\binom{n}{k}} P_{p}( M^{'k}_i) \cdot mIoU_{M'_i}.
    \label{Eua3}
\end{equation}

\begin{table}[h]
\centering
\caption{All combinations of missing modalities and their probabilities under the R-D-E-L modality combination.}
\vspace{-6pt}
\label{tab:modality_probabilities}
\renewcommand{\tabcolsep}{12pt}
\resizebox{0.8\linewidth}{!}{
\begin{tabular}{c|c}
\toprule
\textbf{Modality Combination} & \textbf{Probability \( P \)} \\ \midrule
RGB-Depth-Event-LiDAR            & $(1-p)^4$\\ 
RGB-Depth-Event            & $p(1-p)^3$                    \\ 
RGB-Depth-LiDAR            & $p(1-p)^3$                   \\ 
RGB-Event-LiDAR          & $p(1-p)^3$                \\ 
Depth-Event-LiDAR              & $p(1-p)^3$                   \\ 
RGB-Depth         & $p^2(1-p)^2$                \\ 
RGB-Event        & $p^2(1-p)^2$                \\ 
RGB-LiDAR      & $p^2(1-p)^2$                \\ 
Depth-Event     & $p^2(1-p)^2$                \\
Depth-LiDAR      & $p^2(1-p)^2$               \\
Event-LiDAR            &$p^2(1-p)^2$                    \\ 
RGB           &$p^3(1-p)$                   \\
Depth            & $p^3(1-p)$                    \\
Event            & $p^3(1-p)$                    \\
LiDAR            & $p^3(1-p)$                    \\
\bottomrule
\end{tabular}}
\vspace{-6pt}
\end{table}

\begin{table*}[!t]
    \centering
    \caption{Comparison of MMSS methods under EMM condition of different missing modality combinations (RD means RGB and Depth are the normal modalities).}
    \vspace{-6pt}
    \label{EMMcomparison}
    \small
    \renewcommand{\tabcolsep}{4pt}
    \resizebox{1\linewidth}{!}{
    \begin{tabular}{l|c|ccccccccccccccc}
    \toprule
    \textbf{Model} &  \textbf{Backbone} & \textbf{R} & \textbf{D} & \textbf{E} &\textbf{L} &\textbf{RD} &\textbf{RE}&\textbf{RL}&\textbf{DE}&\textbf{DL}&\textbf{EL}&\textbf{RDE}&\textbf{RDL}&\textbf{REL} &\textbf{DEL}&\textbf{RDEL}\\ \midrule

    CMNeXt~\cite{zhang2023delivering}  & MiT-B2 &22.50 & 50.59&3.16 & 2.86&66.33 &22.92 &22.50 &50.80 & 50.83& 3.15& 66.27 &66.38 &22.92 &50.98 & 66.33\\
    GeminiFusion~\cite{jiageminifusion}  & MiT-B2&15.89 & 54.73& 1.70&1.70 &66.93 &16.24 & 15.8& 54.83& 54.76& 1.7&66.92 & 66.93& 16.18&54.86 &66.92 \\

    MAGIC~\cite{zheng2024centering}  & MiT-B2& 42.72&58.39 &1.90 & 1.62&66.10 &42.79 &42.72 &58.44 &58.39 &1.90 &66.11 & 66.10& 42.80& 58.44&66.10 \\
    MAGIC++~\cite{zheng2024magic++}  & MiT-B2&41.10 & 58.12&2.14 & 1.64& 67.33&41.13 &41.13 &58.32 &58.12 &2.15 &67.35 &67.33 &41.17 & 58.31& 67.34\\
    StitchFusion~\cite{li2024stitchfusion}  & MiT-B2& 30.93& 55.44&1.87 & 1.59&68.22 &31.03 &33.55 & 55.76&55.41& 1.87& 68.23& 68.21&33.66 &55.71 &68.20\\
    \bottomrule
    \end{tabular}
    }
    \vspace{-6pt}
\end{table*}

\begin{table*}[!t]
    \centering
    \caption{Comparison of MMSS methods under RMM condition of different missing modality combinations ($r=0.75$)(RD means RGB and Depth are the normal modalities).}
    \vspace{-6pt}
    \label{RMMcomparison1}
    \small
    \renewcommand{\tabcolsep}{4pt}
    \resizebox{1\linewidth}{!}{
    \begin{tabular}{l|c|ccccccccccccccc}
    \toprule
    \textbf{Model} &  \textbf{Backbone} & \textbf{R} & \textbf{D} & \textbf{E} &\textbf{L} &\textbf{RD} &\textbf{RE}&\textbf{RL}&\textbf{DE}&\textbf{DL}&\textbf{EL}&\textbf{RDE}&\textbf{RDL}&\textbf{REL} &\textbf{DEL}&\textbf{RDEL}\\ \midrule

    CMNeXt~\cite{zhang2023delivering}  & MiT-B2&30.05 & 56.34& 7.04& 7.16& 66.26& 29.98& 30.12& 56.54& 56.39&7.07 & 66.28& 66.32& 30.06& 56.61&66.33\\
    GeminiFusion~\cite{jiageminifusion}  &MiT-B2 & 22.56& 58.1& 2.34&2.34 &66.99 &22.27 &22.57 & 58.05&58.08 & 2.34& 66.98&66.94 &22.25 &58.03 &66.92 \\
    MAGIC~\cite{zheng2024centering}  & MiT-B2& 42.77&58.70 &3.03 &2.83 &66.10 &42.85 &42.77 &58.76 &58.70 &3.02 & 66.10& 66.10&42.86 & 58.76&66.10 \\
    MAGIC++~\cite{zheng2024magic++}  & MiT-B2& 41.22& 59.97&10.52 &10.30 &67.33 & 41.20&41.25 &60.15 &59.97 &10.59 &67.34 &67.33 & 41.23& 60.15& 67.34\\
    StitchFusion~\cite{li2024stitchfusion}  &MiT-B2 &37.18 & 57.79&6.71 &7.18 &68.24 &37.26 &38.51 &58.01 &57.91 & 7.27& 68.23&68.21 &38.57 &58.12 &68.21\\
    \bottomrule
    \end{tabular}
    }
    \vspace{-6pt}
\end{table*}

\begin{table*}[!t]
    \centering
    \caption{Comparison of MMSS methods under RMM condition of different missing modality combinations ($r=0.5$).}
    \vspace{-6pt}
    \label{RMMcomparison2}
    \small
    \renewcommand{\tabcolsep}{4pt}
    \resizebox{1\linewidth}{!}{
    \begin{tabular}{l|c|ccccccccccccccc}
    \toprule
    \textbf{Model} &  \textbf{Backbone} & \textbf{R} & \textbf{D} & \textbf{E} &\textbf{L} &\textbf{RD} &\textbf{RE}&\textbf{RL}&\textbf{DE}&\textbf{DL}&\textbf{EL}&\textbf{RDE}&\textbf{RDL}&\textbf{REL} &\textbf{DEL}&\textbf{RDEL}\\ \midrule

    CMNeXt~\cite{zhang2023delivering}  & MiT-B2&38.26 &58.94& 19.26 &19.32 & 66.29& 38.26& 38.26&59.03 &59.09 &19.32 & 66.30& 66.32& 38.26&59.04 &66.33\\
    GeminiFusion~\cite{jiageminifusion}  & MiT-B2& 29.33& 59.48& 4.38& 4.41& 67.01& 29.22& 29.37& 59.53&59.55 & 4.35& 66.98& 66.93& 29.21&59.48 &66.92 \\
    MAGIC~\cite{zheng2024centering}  & MiT-B2&42.90 & 60.55&14.87 &14.78 & 66.10&42.94 &42.91 &60.58 &60.53 &14.81 &66.10 &66.10 & 42.95&60.57 &66.10 \\
    MAGIC++~\cite{zheng2024magic++}  & MiT-B2&42.13 & 61.54&18.46 &18.51 &67.33 &42.1 &42.17 &61.63 &61.52 & 18.45& 67.34&67.33 & 42.13&61.60 &67.34 \\
    StitchFusion~\cite{li2024stitchfusion}  &MiT-B2 & 41.21&59.46 &15.78 &15.80 &68.22 &41.28 &41.96 & 59.62&59.52 & 15.89&68.21 &68.20 &41.97 & 59.60&68.19\\
    \bottomrule
    \end{tabular}
    }
    \vspace{-6pt}
\end{table*}

\begin{table*}[!t]
    \centering
    \caption{Comparison of MMSS methods under RMM condition of different missing modality combinations ($r=0.25$)}
    \vspace{-6pt}
    \label{RMMcomparison3}
    \small
    \renewcommand{\tabcolsep}{4pt}
    \resizebox{1\linewidth}{!}{
    \begin{tabular}{l|c|ccccccccccccccc}
    \toprule
    \textbf{Model} &  \textbf{Backbone} & \textbf{R} & \textbf{D} & \textbf{E} &\textbf{L} &\textbf{RD} &\textbf{RE}&\textbf{RL}&\textbf{DE}&\textbf{DL}&\textbf{EL}&\textbf{RDE}&\textbf{RDL}&\textbf{REL} &\textbf{DEL}&\textbf{RDEL}\\ \midrule

    CMNeXt~\cite{zhang2023delivering}  & MiT-B2&47.71 &60.92 &34.61 &34.67 &66.31 &47.78 &47.75 &60.94 &61.01 &34.73 & 66.31& 66.32& 47.77& 60.98&66.33\\
    GeminiFusion~\cite{jiageminifusion}  & MiT-B2&44.45 & 61.2& 18.61& 18.66&66.98 & 44.41&44.42 &61.28 &61.21 &18.46 &66.98 &66.93 &44.41 &61.24 &66.92 \\
    MAGIC~\cite{zheng2024centering}  &MiT-B2 &43.96 & 61.98&28.62 & 28.62&66.10 &43.97 & 43.96&62.04 & 61.99&28.68 &66.10 &66.10 &43.96 & 62.01&66.10 \\
    MAGIC++~\cite{zheng2024magic++}  &MiT-B2 & 47.06&62.83 &33.23 &33.16 &67.33 &47.06 &47.07 & 62.88&62.83 &33.30 &67.34 & 67.33&47.10 & 62.89& 67.34\\
    StitchFusion~\cite{li2024stitchfusion}  &MiT-B2 &47.19 &61.47 &29.26 &29.44 &68.24 &47.13 &47.38 &61.53 &61.55 &29.48 &68.24 &68.25 &47.45 &61.61 &68.25\\
    \bottomrule
    \end{tabular}
    }
    \vspace{-12pt}
\end{table*}

\subsubsection{Random-Missing Modality}
Except for sensor damage, in real-world applications, sensor data can be partially missing due to temporary obstructions, noise, or other random factors. In this scenario, a certain proportion of each modality’s data is zeroed out, rather than completely missing a modality. Such a condition reasonably models the unpredictability of sensor failures.

To evaluate the model's performance under random missing modalities, similar to Section \ref{EMM}, we define \( mIoU^{Avg}_{RMM} \) and \( mIoU^{E}_{RMM} \) to measure the model's ability with random missing modality.
For each modality \( m_i \), a proportion \( r \) of the data is randomly set to zero, where \( r \) reflects the proportion of missing data for modality \( m_i \). We define the random-missing modality combinations as \( \{ M_1'', M_2'', \dots, M_N'' \} \subseteq M \), where each \( M''_j \) is a combination that is made up of the modalities partly zeroed.
\( mIoU^{Avg}_{RMM} \) and \( mIoU^{E}_{RMM} \) are calculated as Eq.~\ref{Eq:RmAvg} and \ref{Eq:RmE}.
\( mIoU_{M''_j}(r_j) \) represents the validation mIoU for modality combination \( M''_j \) with \( r \) proportion of its data missing, using model weights derived from training with the complete modality combination. When validating, the modality that is not included in $M''_j$ remains unchanged. In Eq.~\ref{Eq:RmE}, $p$ refers to the probability of each modality for random-missing, which also follows a Bernoulli distribution as Eq.~\ref{Eua2}.
\begin{equation}
    mIoU_{RMM}^{Avg} = \frac{1}{N} \sum_{j=1}^{N} mIoU_{M''_j}(r),
    \label{Eq:RmAvg}
\end{equation}
\begin{equation}
mIoU^{E}_{RMM}(p) = \sum_{k=0}^{n-1} \sum_{j=1}^{\binom{n}{k}} P_{p}( M^{''k}_j) \cdot mIoU_{M''_j}.
    \label{Eq:RmE}
\end{equation}

\subsubsection{Noisy Modality}
In real-world applications, noise interference is inevitable. Thus, evaluating model robustness under noisy conditions is essential, which can help to ensure its reliability in practical applications.
To simulate the real world, we introduce 2 common noises: Gaussian noise and salt-and-pepper noise.

Gaussian noise $N_G$ is used to simulate electronic sensor noise (e.g., CMOS thermal noise) and random disturbances during transmission. Because it is global, continuous, and smooth, Gaussian noise provides a realistic approximation of noise in real-world images evenly distributed across all pixels of the image, resulting in a smooth degradation of image quality. The probability density function of $N_G$  is shown in Eq.~\ref{Eq:G}, which is determined by $\sigma$ and $\mu$.
\begin{equation}
f(x) = \frac{1}{\sigma \sqrt{2\pi}} e^{-\frac{(x - \mu)^2}{2\sigma^2}}.
\label{Eq:G}
\end{equation}

Salt-And-Pepper noise $N_{SP}$  is used to simulate scenarios such as sensor dead pixels, data transmission errors, and dust occlusion. Unlike Gaussian noise, salt-And-Pepper noise is local, discrete, and sharp, effectively reflecting the impact of these issues on the image. $N_{SP} $ is characterized by random extreme pixels appearing in the image, usually black (pepper) and white (salt), which makes the image visually distorted. In validation, the noisy density of $N_{SP}$ is defined as $D$.
During validation, some adjustments are applied to better simulate the real world: 1. Considering that the event stream is asynchronous, there will be no $N_G$ in the event data in practical applications. Thus, to event data, $N_G$ is not applied. 2. Considering that the commonly used definition of $N_{SP}$ noise is based on RGB data, the black and white values in $N_{SP}$ are defined as the maximum and minimum values for the data of other modalities.
We define $mIoU_{NM}$ to evaluate the model's ability under noisy modalities, which is the validation mIoU with model weights derived from training using the noise-free modality. With the origin input as $X$, the noisy input $X_N$ for $mIoU_{NM}$ is defined as Eq.~\ref{Eq:XN}.
\begin{equation}
X_N = X + N_{G}(\sigma, \mu) + N_{SP}(D).
\label{Eq:XN}
\end{equation}
\section{Experiments}

\subsection{Experimental Details}

\begin{table}[!t]
    \centering
    \caption{EMM evaluation results.}
    \vspace{-6pt}
    \label{EMMMetrix}
    \small
    \renewcommand{\tabcolsep}{1pt}
    \resizebox{1\linewidth}{!}{
    \begin{tabular}{l|cccc}
    \toprule
    \textbf{Model} & \textbf{$mIoU^{Avg}_{EMM}$} & \makecell{\textbf{$mIoU^{E}_{EMM}$}\\ \textbf{$(p=0.2)$}} & \makecell{\textbf{$mIoU^{E}_{EMM}$}\\ \textbf{$(p=0.1)$}} &\makecell{\textbf{$mIoU^{E}_{EMM}$}\\ \textbf{$(p=0.05)$}}\\ \midrule
    CMNeXt~\cite{zhang2023delivering} & 37.90 & 54.46 & 60.41 & 63.38 \\
    GeminiFusion~\cite{jiageminifusion} & 37.07 & 54.33 & 60.62 & 63.77 \\
    MAGIC~\cite{zheng2024centering} & 44.97 & 58.66 & 62.68 & 64.47 \\
    MAGIC++~\cite{zheng2024magic++} & 44.85 & 59.18 & 63.52 & 65.50 \\
    StitchFusion~\cite{li2024stitchfusion} & 41.98 & 58.02 & 63.29 & 65.80 \\
    \bottomrule
    \end{tabular}
    }
    \vspace{-6pt}
\end{table}

\begin{table}[!t]
    \centering
    \caption{RMM evaluation results ($r=0.75$).}
    \vspace{-6pt}
    \label{RMMMetrix1}
    \small
    \renewcommand{\tabcolsep}{1pt}
    \resizebox{1\linewidth}{!}{
    \begin{tabular}{l|cccc}
    \toprule
    \textbf{Model} & \textbf{$mIoU^{Avg}_{RMM}$} & \makecell{\textbf{$mIoU^{E}_{RMM}$}\\ \textbf{$(p=0.2)$}} & \makecell{\textbf{$mIoU^{E}_{RMM}$}\\ \textbf{$(p=0.1)$}} &\makecell{\textbf{$mIoU^{E}_{RMM}$}\\ \textbf{$(p=0.05)$}}\\ \midrule
    CMNeXt~\cite{zhang2023delivering} & 42.17 & 56.66 & 61.60 & 63.99 \\
    GeminiFusion~\cite{jiageminifusion} & 39.78 & 55.88 & 61.47 & 64.22 \\
    MAGIC~\cite{zheng2024centering} & 45.30 & 58.77 & 62.72 & 64.49 \\
    MAGIC++~\cite{zheng2024magic++} & 47.06 & 59.81 & 63.78 & 65.62 \\
    StitchFusion~\cite{li2024stitchfusion} & 45.16 & 59.44 & 64.02 & 66.17 \\
    \bottomrule
    \end{tabular}
    }
    \vspace{-12pt}
\end{table}

\begin{table}[!t]
    \centering
    \caption{RMM evaluation results ($r=0.5$).}
    \vspace{-6pt}
    \label{RMMMetrix2}
    \small
    \renewcommand{\tabcolsep}{1pt}
    \resizebox{1\linewidth}{!}{
    \begin{tabular}{l|cccc}
    \toprule
    \textbf{Model} & \textbf{$mIoU^{Avg}_{RMM}$} & \makecell{\textbf{$mIoU^{E}_{RMM}$}\\ \textbf{$(p=0.2)$}} & \makecell{\textbf{$mIoU^{E}_{RMM}$}\\ \textbf{$(p=0.1)$}} &\makecell{\textbf{$mIoU^{E}_{RMM}$}\\ \textbf{$(p=0.05)$}}\\ \midrule
    CMNeXt~\cite{zhang2023delivering} & 47.49 & 58.85 & 62.68 & 64.53 \\
    GeminiFusion~\cite{jiageminifusion} & 42.41 & 57.30 & 62.25 & 64.62 \\
    MAGIC~\cite{zheng2024centering} & 48.19 & 59.53 & 63.01 & 64.61 \\
    MAGIC++~\cite{zheng2024magic++} & 49.31 & 60.50 & 64.07 & 65.75 \\
    StitchFusion~\cite{li2024stitchfusion} & 48.33 & 60.58 & 64.53 & 66.41 \\
    \bottomrule
    \end{tabular}
    }
    \vspace{-12pt}
\end{table}

\begin{table}[!t]
    \centering
    \caption{RMM evaluation results ($r=0.25$).}
    \vspace{-6pt}
    \label{RMMMetrix3}
    \small
    \renewcommand{\tabcolsep}{1pt}
    \resizebox{1\linewidth}{!}{
    \begin{tabular}{l|cccc}
    \toprule
    \textbf{Model} & \textbf{$mIoU^{Avg}_{RMM}$} & \makecell{\textbf{$mIoU^{E}_{RMM}$}\\ \textbf{$(p=0.2)$}} & \makecell{\textbf{$mIoU^{E}_{RMM}$}\\ \textbf{$(p=0.1)$}} &\makecell{\textbf{$mIoU^{E}_{RMM}$}\\ \textbf{$(p=0.05)$}}\\ \midrule
    CMNeXt~\cite{zhang2023delivering} & 53.61 & 61.28 & 63.86 & 65.11 \\
    GeminiFusion~\cite{jiageminifusion} & 49.74 & 60.55 & 63.91 & 65.46 \\
    MAGIC~\cite{zheng2024centering} & 51.61 & 60.46 & 60.37 & 64.76 \\
    MAGIC++~\cite{zheng2024magic++} & 53.92 & 62.07 & 64.78 & 66.08 \\
    StitchFusion~\cite{li2024stitchfusion} & 53.10 & 62.34 & 65.39 & 66.85 \\
    \bottomrule
    \end{tabular}
    }
    \vspace{-12pt}
\end{table}

\begin{table*}[t]
\centering
\caption{NM evaluation results.}
\vspace{-6pt}
\setlength{\tabcolsep}{5.6pt}
\renewcommand{\arraystretch}{1.4}
\resizebox{0.9\linewidth}{!}{
\begin{tabular}{cccccccccccccccc}
\toprule
    \multirow{2}{*}{\textbf{Class}}    & \multicolumn{3}{c}{\textbf{CMNeXt~\cite{zhang2023delivering}}} & \multicolumn{3}{c}{\textbf{GeminiFusion~\cite{jiageminifusion}}} & \multicolumn{3}{c}{\textbf{MAGIC~\cite{zheng2024centering}}} &  \multicolumn{3}{c}{\textbf{MAGIC++~\cite{zheng2024magic++}}} & \multicolumn{3}{c}{\textbf{StitchFusion~\cite{li2024stitchfusion}}}   \\
    \cline{2-16} 
    &  \textbf{Low} & \textbf{Mid.} &  \textbf{High} & \textbf{Low} & \textbf{Mid.} &  \textbf{High} & \textbf{Low} & \textbf{Mid.} &  \textbf{High}&\textbf{Low} & \textbf{Mid.} &  \textbf{High}&\textbf{Low} & \textbf{Mid.} &  \textbf{High} \\ \midrule

Building & 54.46 &6.39 &0&0 &0&0&45.88&28.22&9.58 &59.04&41.54&16.42 &51.72&26.35&3.84     \\ 
 Fence& 12.11 & 4.88&1.27&0.21 &0&0 &13.60 & 4.55&0.03&5.85& 1.53&0.75 &12.48&7.90&  1.91  \\ 
 Other& 0 &0 &0& 0&0&0 &0 &0  & 0&0&0& 0&0&0&  0 \\ 
  Pedestrian&32.85&14.10&0.07& 0.14&0&0 &16.38 &6.99&0.12&24.32&16.54 &6.15 &17.38&9.43& 2.17\\
 Pole&  45.34& 24.70&4.29&6.74 &0.21&0 &12.60&2.70&0.02&26.05  &13.46 & 5.06 &14.84&6.17&  1.32\\
 RoadLine& 43.74 &18.88 &1.25&0.01 &0&0 &54.38 &37.39&15.12& 62.53  &48.58 &30.57&48.90&35.49&  22.21\\
 Road& 87.94 &56.83 &3.00&2.54 &0&0 & 75.74 &61.59&49.11& 91.85 & 78.38&68.66&79.10&68.19& 51.83 \\
 SideWalk&  52.54& 27.61&5.11& 0.23&0&0 &39.26 &14.25&0.11&47.56 &32.43 &8.09 &35.57&24.91&  10.31 \\
 Vegetation& 34.46 & 10.34&5.60& 8.96&8.27&7.48 & 16.91 &12.77&1.21&31.23 &9.89 &10.73 &43.20&32.35& 16.84\\
 Cars & 51.99 &26.13 &0.14& 0.04&0&0 & 48.73&31.42&1.00&53.99  & 26.65&14.37 &41.69&31.69& 21.36\\
 Wall & 4.72 & 0.15&0.01&0 &0&0 &21.49&8.69&0.25& 21.52  & 7.54 &1.00 &10.83&4.80& 2.42\\
 Trafficsign & 27.39 &6.22 &0.04&0.01 &0&0 &9.94 &5.90&2.31&15.88 &6.84& 0.60 &10.31&4.33& 0.70\\
 Sky &  95.43&47.66 &7.81&97.65 &94.17&74.96 &41.01  &17.97&10.22&75.31 &9.84&1.18 &89.21&76.61& 40.11\\
Ground&  2.17&0.39 &0&0 &0&0 &1.75  &0.80&0.00& 0.64 & 1.21&0.69&0.74&0.76&0.21 \\
 Bridge & 4.03 &0.23 &0&0 &0&0 &10.24 &2.37&0.01& 36.19 &24.40&1.70 &11.07&2.50& 2.15 \\
RailTrack & 20.78 &0.39 &0&0 &0& 0& 0.78 &0.41&0.21&24.51 &6.80&0.40 &10.81&3.36& 1.47\\
GroundRail & 9.39 &2.08 &0.13&0.04 &0&0 &8.52&2.67&0.07& 13.28  & 7.88&4.26 &9.26&4.01& 0.88\\
TrafficLight&  48.61& 25.85&2.93& 3.42&0.01&0 &25.87 &8.20&0.90& 46.87 &24.93& 3.52&27.99&9.90& 1.60 \\
Static& 14.51 &1.94 &0.01& 0&0&0 & 8.58&2.58&0.05&12.3 &5.47& 2.02 &8.35&3.72&1.16 \\
Dynamic &  5.71&0.64 &0&0 &0&0 &2.96&0.94&0&  5.55 &2.30&0.15 &3.87&0.94& 0\\
Water& 33.43 & 19.92&2.13&6.35 &0&0 &5.14&2.97&0.34& 3.90 &1.79& 0.49 &0.05&0.01& 0 \\
Terrain& 48.80 & 26.52&13.12&10.06 &0.88&0 &29.26 &17.43&2.64& 42.75  &22.17&6.88&33.76&20.54& 5.45 \\
TwoWheeler & 17.54 &5.40 &0&0 &0&0 &17.22 &5.93&0& 14.00  & 6.86&2.16&16.78&10.82& 1.49\\
Bus &  60.42& 32.25&0.39& 0.18&0&0 & 37.79&28.11&5.93& 53.28  &31.44 &20.52&36.69&38.30& 38.37\\
 Truck& 72.42 &49.64 &10.45& 1.49&0&0 & 56.60&27.94&0.30& 59.52 &29.19 &11.01 &8.20&4.76& 3.40\\ \hline
 $mIoU_{NM}$ & 35.23 &16.37 &2.31&5.52 &4.14&3.30&24.03&13.31&3.98&33.12&18.31&8.70&24.91 & 17.11 &9.25 \\

  \bottomrule

\end{tabular}
}
\label{NM}
\vspace{-8pt}
\end{table*}

All experiments are conducted on 1 A800 GPU with a batch size of 4. Same as~\cite{zhang2023delivering}, we validate the models with the image size set $1024\times 1024$. To verify the conjectures in Section~\ref{MMSS} and ensure the comprehensiveness of the experiment, we select the representative methods of the three types of MMSS methods mentioned in Section~\ref{MMSS} for validation. In more detail, we choose CMNeXt~\cite{zhang2023delivering} for Type\textcircled{1}, MAGIC~\cite{zheng2024centering}, MAGIC++~\cite{zheng2024magic++} for Type\textcircled{2}, and GeminiFusion~\cite{jiageminifusion}, StitchFusion~\cite{li2024stitchfusion} for Type\textcircled{3}. For a fair comparison, MiT-B2~\cite{xie2021segformer} is selected as the backbone.

For the Entire-Missing Modality (EMM) condition, we first validate the mIoU of each possible combination for the subsequent calculations. The calculation of $mIoU^{Avg}_{EMM}$ follows Eq.~\ref{Eua1}. For $mIoU^{E}_{EMM}$, we set $p$ to 3 different values: 0.2, 0.1, and 0.05, representing the different frequencies at which the damage occurs.
For the Random-Missing Modality (RMM) condition, the experimental settings are the same as EMM, with $p'$ set to the same values as $p$. However, when validating in the RMM condition, another hyper-parameter $r$ exists, which reflects the proportion of data missing. For the sake of experimental completeness, we set up three degrees of data missing based on the value of $r$. In more detail, $r=0.75$ refers to a high level of data missing, $r=0.5$ refers to a middle level of data missing, and $r=0.25$ refers to a low level of data missing.
For the Noisy Modality (NM) condition, we validate the model on the RGB-Depth-Event-LiDAR modality combination. Similar to RMM, we set up three degrees of data noise based on the values of $D$ and $\sigma$. $D=0.2, \sigma = 0.5$ refers to a high level of data missing, $D=0.1, \sigma = 0.2$ refers to a middle level of data missing, and $D=0.05, \sigma = 0.1$ refers to a low level of data missing.
The $\mu$ is set to 0. Expected values are additionally normalized at the end to achieve a fair comparison of indicators.
\subsection{Experimental Results}
\label{experiments}
\subsubsection{Entire-Missing Modality}
\begin{figure}[t!]
    \centering
    \includegraphics[width=0.85\linewidth]{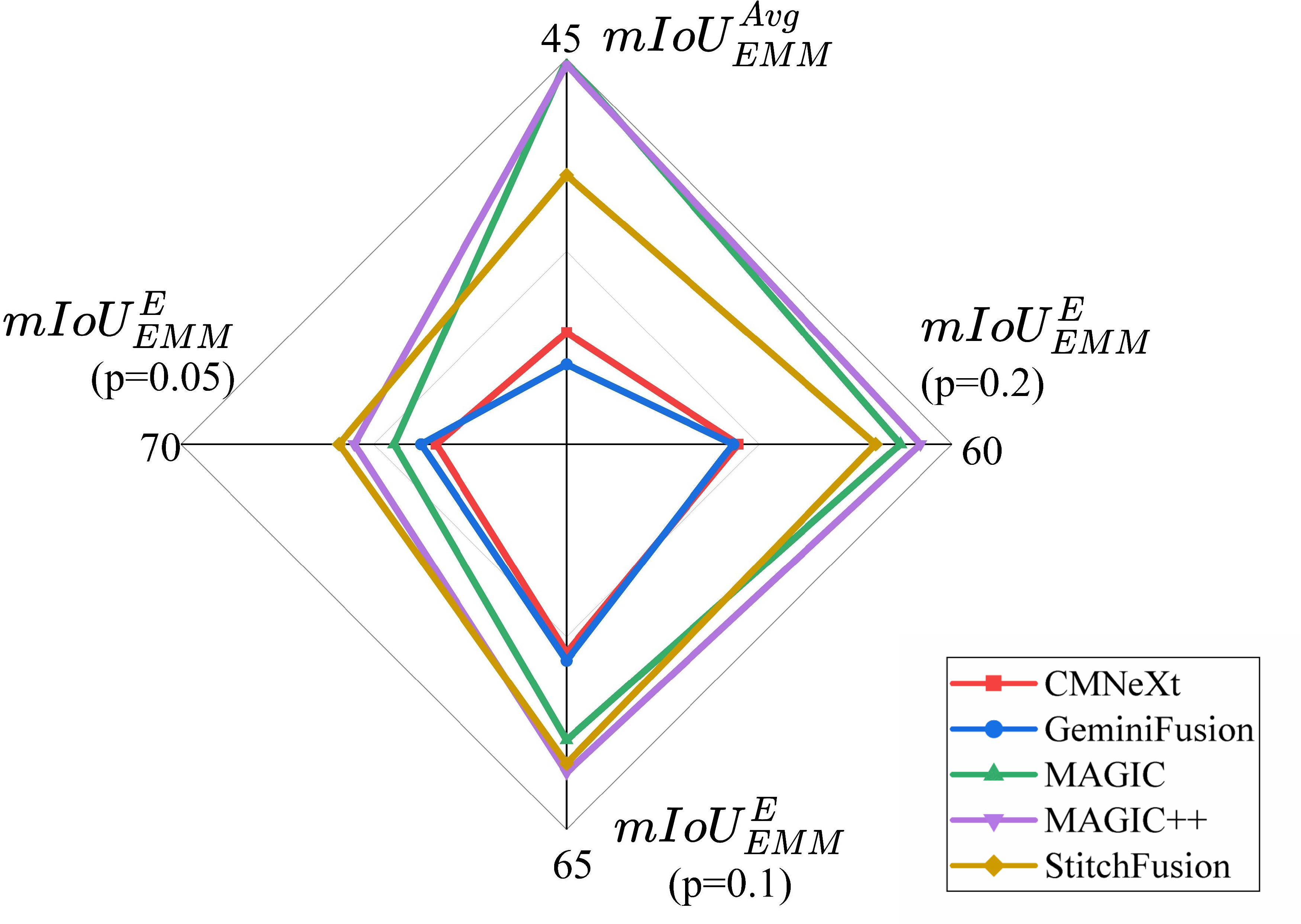}
    \vspace{-3pt}
    \caption{Visualization of EMM validation results. The scale of the radar chart is set to 5.}
    \label{fig:EMM}
\end{figure}

The results of EMM validation are shown in Table~\ref{EMMcomparison}, \ref{EMMMetrix} and Fig.~\ref{fig:EMM}.
In the EMM condition, some of modalities are dropped, allowing a direct validation of each modality's contribution and models' robustness. Firstly, as shown in Table~\ref{EMMcomparison}, the CMNeXt meets a huge decrease when the RGB modality is gone. For example, when it comes to DE, DL, and DEL combinations, the mIoU goes down by 15.53\%, 15.50\%, and 15.35\%, which are the most significant decreases among the 5 models. Such results prove that Type\textcircled{1} models rely too much on the RGB modality, which causes the potential risk when RGB cameras are broken or disturbed. Thus, the first conjecture in Section \ref{MMSS} can be initially proved. 
Secondly, even if the MAGIC and MAGIC++ don't achieve the greatest mIoU under the RDEL combination, these 2 Type\textcircled{3} models show a surprising robustness in the EMM condition. MAGIC++ achieves the best $mIoU^{Avg}_{EMM}$, $mIoU^{E}_{EMM} (p=0.2)$, and $mIoU^{E}_{EMM} (p=0.1)$, while MAGIC achieves the second best $mIoU^{Avg}_{EMM}$ and $mIoU^{E}_{EMM} (p=0.2)$. Thus, it can be proved that the adaptive selection methods in Type\textcircled{3} models effectively compress the fail modality's influence on the entire model, which is the third conjecture in Section \ref{MMSS}.
Thirdly, as to the Type\textcircled{2} models, GeminiFusion achieves a similar ability to CMNeXt while StitchFusion surpasses CMNeXt but still falls behind MAGIC and MAGIC++. The fusion strategy of Type\textcircled{2} models is effective in avoiding too much reliance on some specific modalities, while it works little to help restrain the failed modality's influence, which is the same as the second conjecture in Section \ref{MMSS}.

\subsubsection{Random-Missing Modality} 
The results of RMM validation are shown in Table~\ref{RMMcomparison1}, \ref{RMMcomparison2}, \ref{RMMcomparison3}, \ref{RMMMetrix1}, \ref{RMMMetrix2}, and \ref{RMMMetrix3}. 
As shown in Table~\ref{RMMcomparison1}, \ref{RMMcomparison2}, and \ref{RMMcomparison3}, StitchFusion performs better in certain modality combinations, especially consistently outperforming MAGIC in RD combinations, indicating that it has an advantage in dealing with specific modality combinations. Meanwhile, MAGIC++ is stable in most modality combinations.
As the proportion of missing decreases, the performance of both methods improves significantly, especially in the case of combinations including more modalities.
StitchFusion has certain advantages when dealing with more modalities, and may be more suitable for scenarios where modality is less lost.
MAGIC++ shows an advantage when the modality is severely lost.

\begin{figure}[t!]
    \centering
    \includegraphics[width=0.85\linewidth]{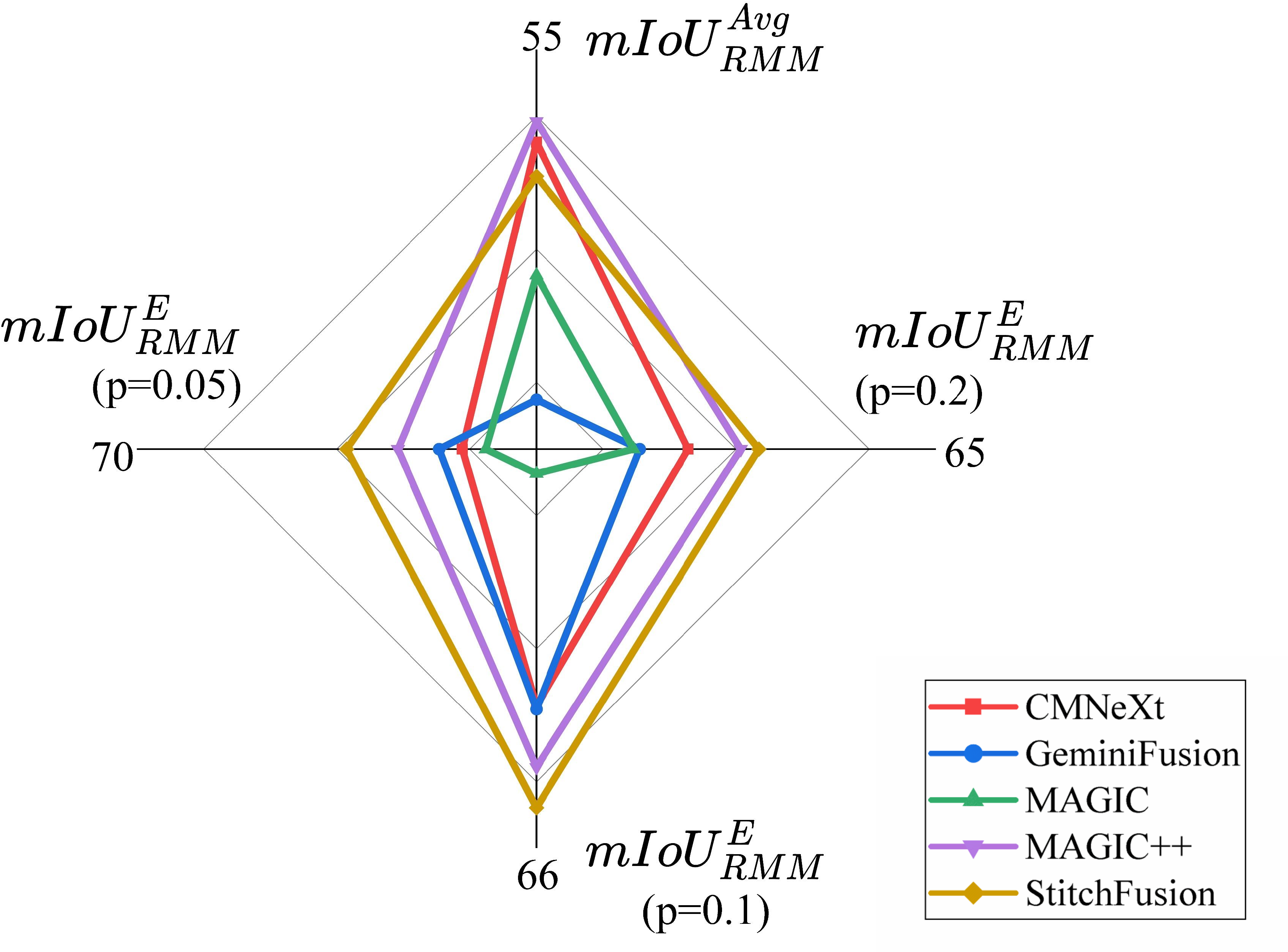}
    \vspace{-3pt}
    \caption{Visualization of RMM validation results with $r=0.25$. The scale of the radar chart is set to 2.}
    \vspace{-8pt}
    \label{fig:RMM}
\end{figure}

As shown in Table~\ref{RMMMetrix1}, \ref{RMMMetrix2}, \ref{RMMMetrix3}, and Fig.~\ref{fig:RMM}, $mIoU_{RMM}^{Avg}$ of MAGIC++ is slightly higher than that of StitchFusion at all proportions of data missing, showing its superiority in the case of RMM.
As the $p$ value decreases, the $mIoU_{RMM}^{E}$ of both models shows an upward trend.
StitchFusion's mIoU growth is slightly larger at low $p$ values, especially when $p=0.05$, StitchFusion's mIoU slightly outperforms MAGIC++.

\subsubsection{Noisy Modality}
As shown in Table~\ref{NM}, the NM validation results differ significantly between different models. CMNeXt's $mIoU_{NM}$ ranks first when the noisy level is low, achieving 35.23\%. However, when the noisy level goes up, $mIoU_{NM}$ goes through the most significant decrease, with 16.37\% at the middle level and 2.31\% at the high level. Such a phenomenon is most likely due to CMNeXt's over-reliance on RGB modality. The RGB modality is a relatively information-intensive and robust modality. Thus, when the noise keeps a low level, the RGB modality is able to correct noise in other modalities. When the RGB modality collapses due to noise, the entire model quickly collapses with it. 
Meanwhile, GeminiFusion shows surprisingly poor performance. We believe that this is due to the excessive inter-modal information exchange in the model architecture, which leads to the continuous propagation of noise in the model, thus expanding the influence of noise on the modality. In addition, the MAGIC++ shows a relatively stable and leading performance in the NM condition. Such a finding once again proves the excellent ability of the Type\textcircled{3} model in terms of MMSS robustness.
\subsection{More Discussions}
\label{Discussion}
As shown in Table~\ref{EMMcomparison}, \ref{RMMcomparison1}, \ref{RMMcomparison2}, and \ref{RMMcomparison3}, Event and LiDAR modalities appear to be relatively redundant. For example, in Table~\ref{EMMcomparison}, the mIoU of RD, RDE, RDL, and RDEL are almost the same. For CMNeXt, the mIoU of REDL is 66.33\%, while the mIoU of RDL is 0.05\% higher. For StitchFusion, the mIoU of RDEL is 68.20\%, while the mIoU of RD is 0.02\% higher. Moreover, the contributions of Depth and LiDAR modalities seem highly similar. For example, in Table~\ref{EMMcomparison}, the mIoU of RE and RL, RDE, and RDL are almost the same across all models.
Such findings suggest that the MMSS model can be optimized by reducing the redundancy between Event and LiDAR modalities. This can be achieved through feature fusion or attention mechanisms, emphasizing the unique contribution of each modality while minimizing overlap.
In addition, understanding the redundancy between Event and LiDAR can guide sensor deployment decisions in real-world applications, potentially reducing costs by selectively using one modality over the other without significant performance loss. 
\section{Conclusion}
In this study, we create a comprehensive benchmark of robustness in Multi-Modal Semantic Segmentation (MMSS), which is essential for the related models' successful deployment in real-world scenarios characterized by uncertain data quality. This benchmark evaluates models under three challenging scenarios: Entire-Missing Modality (EMM), Random-Missing Modality (RMM), and Noisy Modality (NM).
To further enhance the robustness evaluation, we model modality failure events from a probabilistic perspective, considering two key conditions: equal probability for each damaged modality combination and modality damage following a Bernoulli distribution. Based on these assumptions, we have developed four metrics ($mIoU^{Avg}_{EMM}, mIoU^{E}_{EMM}, mIoU^{Avg}_{RMM}, mIoU^E_{RMM}$) to provide a more reasonable assessment of model performance in EMM and RMM scenarios.
Our work represents a pioneering effort to establish a robustness benchmark for MMSS, offering valuable insights and a foundation for future research in this field. By bridging the gap between theory and application, we aim to facilitate the development of more resilient MMSS models that can effectively handle the complexities of real-world multi-modal systems.
\section*{Acknowledgement}
This work was supported by the Guangdong Provincial Department of Education Project (Grant No.2024KQNCX028); CAAI-Ant Group Research Fund; Scientific Research Projects for the Higher-educational Institutions (Grant No.2024312096), Education Bureau of Guangzhou Municipality; Guangzhou-HKUST(GZ) Joint Funding Program (Grant No.2025A03J3957), Education Bureau of Guangzhou Municipality.

\clearpage
{
    \small
    \bibliographystyle{ieeenat_fullname}
    \bibliography{Ref}
}


\end{document}